\documentclass[11pt]{article}
\usepackage{graphicx}
\usepackage[final]{acl}

\usepackage{times}
\usepackage{latexsym}
\usepackage[T1]{fontenc}
\usepackage[utf8]{inputenc}
\usepackage{microtype}
\usepackage{inconsolata}
\usepackage{graphicx}
\usepackage{array} 
\usepackage{cuted}
\usepackage{capt-of}
\usepackage{enumitem}
\title{A Picture is Worth a Thousand Words? An Empirical Study of Aggregation Strategies for Visual Financial Document Retrieval}

\author{Ho Hung Lim and Yi Yang\\
  The Hong Kong University of Science and Technology \\
  \texttt{limhohung@ust.hk}, \texttt{imyiyang@ust.hk} \\}

\begin{document}
\maketitle


\begin{abstract}
Visual RAG has offered an alternative to traditional RAG. It treats documents as images and uses vision encoders to obtain vision patch tokens. However, hundreds of patch tokens per document create retrieval and storage challenges in a vector database. Practical deployment requires aggregating them into a single vector. This raises a critical question: does single-vector aggregation lose key information in financial documents? We develop a diagnostic benchmark using financial documents where changes in single digits can lead to significant semantic shifts. Our experiments show that single-vector aggregation collapses different documents with almost identical vectors. Metrics show that the patch level detects semantic changes, and confirm that aggregation obscures these details. We identify global texture dominance as the root cause. Our findings are consistent across model scales, retrieval-optimized embeddings, and multiple mitigation strategies, highlighting significant risks for single-vector visual document retrieval in financial applications.
\end{abstract}
\section{Introduction}
\begin{figure}[t]
    \centering
\includegraphics[width=1.1\columnwidth]{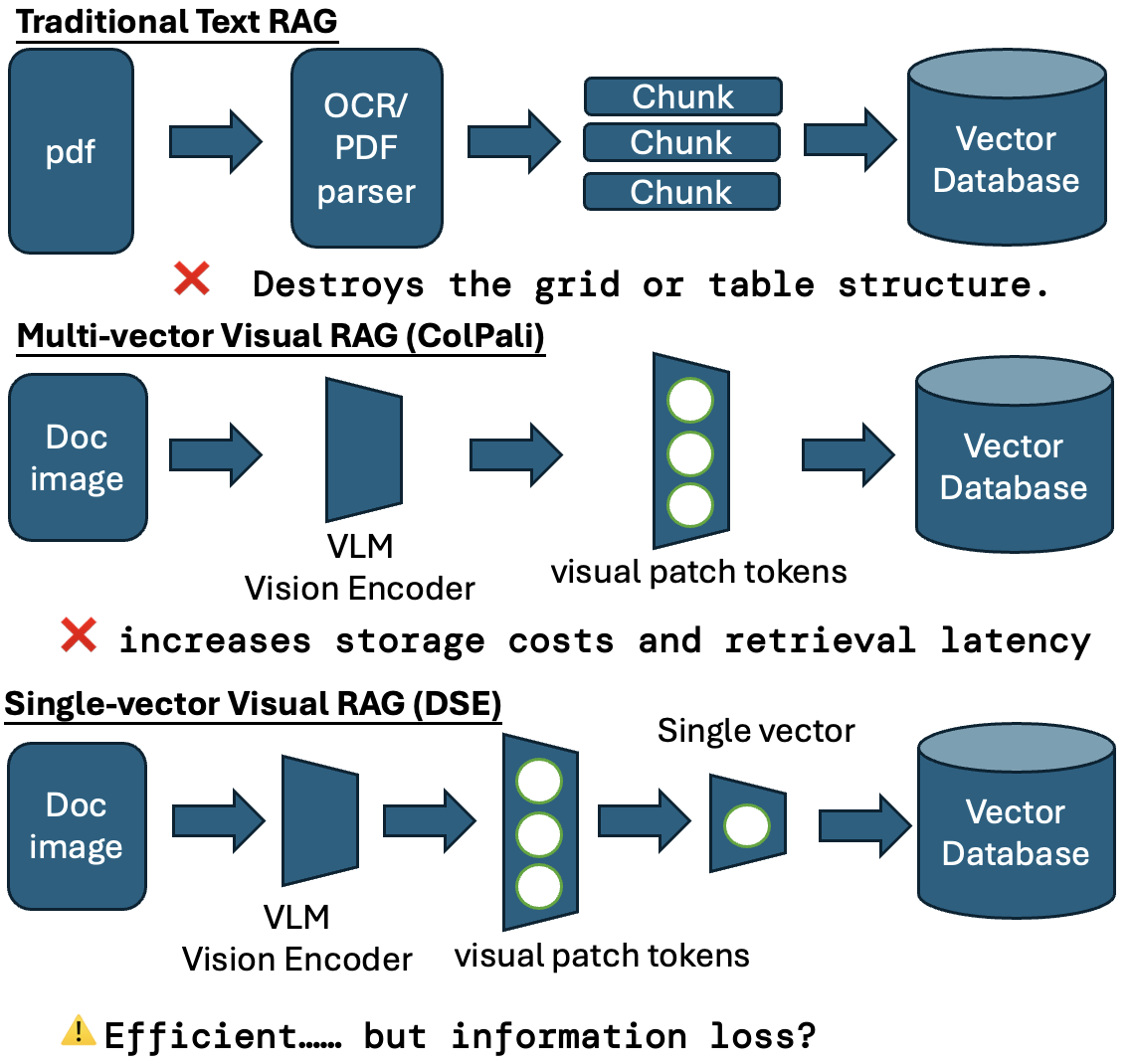}
    \caption{Overview of document retrieval paradigms.}
    \label{fig:overview}
    \vspace{-15pt}
\end{figure}

Retrieval-Augmented Generation (RAG) systems~\cite{10.5555/3495724.3496517} are widely used in the financial domain for analyzing complex financial documents such as annual reports and 10-K reports~\cite{dadopoulos2025metadatadrivenretrievalaugmentedgenerationfinancial,kim2025optimizingretrievalstrategiesfinancial}. They commonly use PDF parsing or Optical Character Recognition (OCR) for extracting document elements and converting them into a linear text sequence~\cite{si2025tabragtabulardocumentretrieval,dadopoulos2025metadatadrivenretrievalaugmentedgenerationfinancial}. However, documents with tables displaying rows and columns are forced to form a linear text sequence that destroys the table structure and breaks row‑column alignments. Because document structural context is destroyed during this process, the document retrieval performance decreases~\cite{yu-etal-2025-tablerag}.

Recent advancements leverage the superior image processing capabilities of Vision-Language Models (VLMs) to extract document content for OCR and retrieval~\cite{kim2022donut,ma-etal-2024-unifying,faysse2024colpali,yu2025visrag}. 
DeepSeek-OCR~\cite{wei2025deepseekocrcontextsopticalcompression} proves the efficacy of encoding long textual context directly into visual tokens that can achieve high accuracy in OCR tasks. Similarly, ColPali~\cite{faysse2024colpali} uses VLMs for the representation of document page images as a sequence of visual patch tokens for document retrieval. Despite its success in effectively utilizing document information conveyed by visual patch features, there is consequently higher storage cost due to embedding hundreds of vectors for document page images in the vector database setup; this is a typical characteristic of multi-vector storage in retrieval systems \cite{10.1007/978-3-031-88714-7_22}. For practical deployment, dense retrieval methods such as DSE~\cite{ma-etal-2024-unifying} compress these visual tokens into a single vector for efficient storage and faster document retrieval.

Table-centric financial documents differ fundamentally from generic visual texts in that their most salient information is often encoded in sparse numerical values and key financial entities, rather than in dominant visual structures. This raises an important research question: \textbf{When we compress visual patch tokens from a dense financial document into a single vector through aggregation, do we lose key numeric or semantic information?} We hypothesize that this loss is particularly severe for financial documents, where dense numerical data appear against dominant background layouts. Although a single digit changes the semantic meaning, the visual signal of that digit is tiny. As a result, aggregation strategies often prioritize the large background features and smooth over these sparse numeric details. Therefore, we address two critical questions in the financial domain:
\begin{enumerate}[noitemsep, topsep=2pt]
    \item \textbf{Severity}: How severe is this information loss for financial documents, where a single digit change (e.g., \$1.2M $\to$ \$7.2M) or a textual shift (e.g., changing a date) represents a significant semantic difference?
    \item \textbf{Mechanism}: What causes this failure? 
\end{enumerate}

To answer these questions, we build a diagnostic benchmark. We find that single-vector aggregation fails to preserve the fine-grained details detected by the encoder. We make the following contributions:

\begin{itemize}[noitemsep, topsep=2pt]
\item \textbf{Aggregation Failure.} We show that 
\textbf{single-vector aggregation} causes different document images to look almost identical (similarity 
$>0.99$), consistently across model scales (7B to 32B) and retrieval-optimized embeddings.

\item \textbf{Diagnostic Proof.} We use \textbf{MinPatch} to find the signal. It recovers the score (Similarity $\approx 0.51$). This proves the encoder features are good, but the aggregation ruins them. Simple mitigation strategies also fail to recover the signal.

\item \textbf{Mechanistic Explanation.} We find the cause is \textbf{global texture dominance}. Our analysis shows that single-vector aggregation focuses on the background layout or grid lines instead of the table data.
\end{itemize}

\section{Related Work}
\subsection{VLMs and Compression}
Modern Vision-Language Models (VLMs) like LLaVA~\cite{10.5555/3666122.3667638} and Qwen2.5-VL~\cite{bai2025qwen25vltechnicalreport} encode images into sequences of visual patch tokens through their vision encoder. Recently, DeepSeek-OCR~\cite{wei2025deepseekocrcontextsopticalcompression} achieves remarkable OCR performance by effectively compressing extremely long text contexts into visual token representations. It proves that sequences of visual patch tokens can preserve fine-grained visual details. Although these patch representations are highly effective in generation tasks, they are impractical in the context of retrieval due to the cost of storing hundreds of tokens for every document in a large-scale vector storage system.

\subsection{Visual Document Retrieval}
Existing methods for visual document retrieval are generally of two types: single-vector and multi-vector approaches. Single-vector approaches, like DSE~\cite{ma-etal-2024-unifying}, compress visual tokens into a single vector through aggregation (mean pooling or [CLS] token) for efficient retrieval. Though scalable, they fail to retain details of complex layouts. Multi-vector approaches, represented by ColPali~\cite{faysse2024colpali}, use late interaction (MaxSim) over visual patches. ColPali maintains vision tokens for all patches, retaining fine-grained visual details at the cost of storage and latency overhead. 
\section{Methodology}

\subsection{Sensitivity Analysis}
In order to evaluate whether the vision embeddings accurately capture the semantic meaning within financial document images, we divide our sensitivity analysis into two categories: numeric sensitivity and text sensitivity.

\subsection{Numeric Sensitivity}
We investigate numeric sensitivity using two distinct perturbation levels. 
\begin{table}[h]
\centering
\small
\begin{tabular}{l|l|l}
\hline
\textbf{Condition} & \textbf{Original} & \textbf{Counterfactual} \\
\hline
\textbf{Micro-Semantic} & 5.21 & 5.29 \\
(Small Change) & 19.65\% & 19.54\% \\
 & 10,520 & 10,526 \\
\hline
\textbf{Macro-Semantic} & 5.21 & 9.99 \\
(Large Change) & 11.9 & 88.8 \\
 & 13,499 & 99,999 \\
\hline
\end{tabular}
\caption{Examples of numeric sensitivity strategies.}
\label{tab:micro_macro_examples}
\vspace{-15pt}
\end{table}

\subsection{Text Sensitivity}
We extract the question-answer pairs from the dataset ground truth and manually locate the visual span of the answer within the document image. We then adapt the \textit{semantic occlusion} technique proposed by \citet{zeiler2014visualizing}, covering the answer span with the exact background color.

\begin{table}[h]
\centering
\small
\begin{tabular}{l| >{\raggedright\arraybackslash}p{0.35\columnwidth} | >{\raggedright\arraybackslash}p{0.35\columnwidth}}
\hline
\textbf{Type} & \textbf{Original} & \textbf{Occluded} \\
\hline
\textbf{Revenue} & Revenue increased by \textbf{\$1.4 billion}. & Revenue increased by \texttt{[MASK]}. \\
\hline
\textbf{Date} & Period ending \textbf{Dec 31}. & Period ending \texttt{[MASK]}. \\
\hline
\end{tabular}
\caption{Example of text sensitivity strategies. Ground-truth answers are occluded with the background color (symbolized as \texttt{[MASK]}).}
\label{tab:text_mask_example}
\vspace{-2em} 
\end{table}

\subsection{Structure Analysis}
We conduct a \textbf{visual attention analysis} to test whether vision embeddings focus on the background layout (including non-informative headers) or the tabular data. For this analysis, we generate three image versions.
\begin{itemize}[noitemsep, topsep=5pt]
    \item \textbf{Reference:} The original document image.
    \item \textbf{Signal Image:} We preserve the table content while filling the entire surrounding area (including headers, footers, and margins) with the document's dominant background color. This isolates the table structure and data from any page context.
    \item \textbf{Noise Image:} We fill the internal table area with the identical background color, effectively "erasing" the table. This leaves only the document template (e.g., logos, pagination) visible against a uniform background.
\end{itemize}
We measure the cosine similarity of the signal and noise images against the reference to determine which component dominates the single vector.

\subsection{Retrieval Mechanisms}
To quantify the information loss caused by vector compression, we compare five scoring mechanisms. Let $V_A = \{v_1^A, ..., v_n^A\}$ and $V_B = \{v_1^B, ..., v_n^B\}$ be the patch embedding sequences for the original and counterfactual documents, respectively.

\noindent\textbf{Aggregation:} Aggregating all patches into a single vector before computing similarity:
\begin{itemize}[noitemsep, topsep=0pt]
    \item \textbf{Mean Pooling:} \\
    $S_{mean} = \cos(\frac{1}{n}\sum v_i^A, \frac{1}{n}\sum v_i^B)$. \\
    Averages all patch vectors into a single vector.
    \item \textbf{Max Pooling:} \\
    $S_{max} = \cos(\max(V_A), \max(V_B))$. \\
    Selects the dimension-wise maximum features.
\end{itemize}

\noindent\textbf{Late Interaction:} Computing similarity at the patch level:
\begin{itemize}[noitemsep, topsep=0pt]
    \item \textbf{MaxSim} \cite{10.1145/3397271.3401075}:\\ $S_{ms} = \frac{1}{n} \sum_{i} \max_{j} \cos(v_i^A, v_j^B)$. \\
    Averages the best-match similarity for every patch.
    \item \textbf{MeanPatch:} $S_{mp} = \frac{1}{n} \sum_{i} \cos(v_i^A, v_i^B)$. \\
    Averages the similarity of spatially aligned patches.
    \item \textbf{MinPatch:} $S_{min} = \min_{i} \cos(v_i^A, v_i^B)$. \\
    Identifies the single worst similarity score to isolate local semantic deviations. We use MinPatch as a diagnostic probe, not as a practical retrieval metric.
\end{itemize}

\subsection{Mitigation Strategies}
To explore whether simple modifications to the aggregation process can mitigate global texture dominance, we design three alternative 
aggregation strategies: variance-weighted pooling, 
attention-guided pooling, and top-k patch removal. 
Details are provided in Appendix~\ref{sec:mitigation_details}.

\section{Experiments}
\begin{table}[h]
\small
\begin{tabular}{l|cc|c}
\hline
\textbf{Experiment} & \textbf{FinQA} & \textbf{TAT-DQA} & \textbf{Total} \\
\hline
\multicolumn{4}{l}{\textbf{\textit{Sensitivity Analysis}}} \\
\hline
Micro-Semantic & 100 & 100 & 200 \\
Macro-Semantic & 100 & 100 & 200 \\
Text Sensitivity & 100 & 100 & 200 \\
\hline
\multicolumn{4}{l}{\textbf{\textit{Visual Attention Analysis}}} \\
\hline
Signal (Table Only) & 100 & 100 & 200 \\
Noise (Context Only) & 100 & 100 & 200 \\
\hline
\end{tabular}
\caption{Dataset statistics. We construct a balanced diagnostic set (N=200 pairs) for sensitivity analysis and visual attention analysis.}
\label{tab:master_dataset_stats}
\vspace{-2em} 
\end{table}
\subsection{Dataset Setup}
We test on two financial datasets: \textbf{FinQA}~\cite{chen-etal-2021-finqa} 
and \textbf{TAT-DQA}~\cite{zhu-etal-2021-tat,Zhu_2022}. 
Details on image extraction and masking procedure are in 
Appendix~\ref{sec:dataset_details}.

\subsection{Models}
Following recent work on analyzing VLM information 
loss~\cite{li-etal-2025-lost-embeddings}, we test the vision encoders of five standard VLMs: \textbf{Qwen2.5-VL-7B/32B}~\cite{bai2025qwen25vltechnicalreport}, \textbf{LLaVA-v1.5} ~\cite{10.5555/3666122.3667638}, \textbf{Phi-3.5-Vision} ~\cite{abdin2024phi3technicalreporthighly}, and \textbf{DeepSeek-DeepEncoder}~\cite{wei2025deepseekocrcontextsopticalcompression}. We extract the sequence of visual patches output by the projection layer for each model. To evaluate whether retrieval-specific 
training resolves the aggregation failure, we additionally test two embedding models: \textbf{Qwen3-VL-Embedding-8B}~\cite{qwen3vlembedding} and \textbf{GME-Qwen2-VL-7B-Instruct}~\cite{zhang2024gme}. Full implementation details are provided in Appendix~\ref{sec:appendix_implementation}.

\clearpage 

\begin{strip}
\small
\resizebox{\textwidth}{!}{
\begin{tabular}{l|ccccc|ccccc}
\hline
& \multicolumn{5}{c|}{\textbf{FinQA}} & \multicolumn{5}{c}{\textbf{TAT-DQA}} \\
\hline
\textbf{Model} & \textbf{Mean} & \textbf{Max} & \textbf{MaxSim} & \textbf{MeanP} & \textbf{MinP} & \textbf{Mean} & \textbf{Max} & \textbf{MaxSim} & \textbf{MeanP} & \textbf{MinP} \\
\hline
\multicolumn{11}{c}{\textit{Micro-Semantic Sensitivity (Values close to 1.0 indicate blindness)}} \\
\hline
Qwen2.5-VL-7B & 1.0000 & 0.9996 & 0.9982 & 0.9975 & \textbf{0.7313} & 1.0000 & 1.0000 & 0.9995 & 0.9994 & \textbf{0.6959} \\
Qwen2.5-VL-32B & 0.9999 & 0.9993 & 0.9979 & 0.9969 & \textbf{0.7123} & 1.0000 & 0.9999 & 0.9995 & 0.9994 & \textbf{0.7107} \\
LLaVA-v1.5 & 1.0000 & 1.0000 & 0.9992 & 0.9974 & \textbf{0.7784} & 1.0000 & 1.0000 & 0.9996 & 0.9991 & \textbf{0.7240} \\
Phi-3.5-Vision & 1.0000 & 0.9999 & 0.9980 & 0.9979 & \textbf{0.8747} & 1.0000 & 1.0000 & 0.9997 & 0.9997 & \textbf{0.9002} \\
DeepEncoder & 0.9999 & 0.9997 & 0.9987 & 0.9984 & \textbf{0.9652} & 1.0000 & 0.9999 & 0.9998 & 0.9998 & \textbf{0.9784} \\
\hline
\multicolumn{11}{c}{\textit{Macro-Semantic Sensitivity}} \\
\hline
Qwen2.5-VL-7B & 0.9998 & 0.9985 & 0.9939 & 0.9906 & \textbf{0.5160} & 0.9998 & 0.9998 & 0.9974 & 0.9955 & \textbf{0.5205} \\
Qwen2.5-VL-32B & 0.9999 & 0.9989 & 0.9936 & 0.9899 & \textbf{0.5289} & 1.0000 & 0.9997 & 0.9977 & 0.9960 & \textbf{0.5635} \\
LLaVA-v1.5 & 0.9999 & 1.0000 & 0.9955 & 0.9883 & \textbf{0.6177} & 0.9999 & 1.0000 & 0.9981 & 0.9947 & \textbf{0.6339} \\
Phi-3.5-Vision & 1.0000 & 0.9989 & 0.9925 & 0.9920 & \textbf{0.7408} & 1.0000 & 0.9994 & 0.9974 & 0.9969 & \textbf{0.7838} \\
DeepEncoder & 0.9995 & 0.9985 & 0.9915 & 0.9882 & \textbf{0.8830} & 0.9997 & 0.9997 & 0.9976 & 0.9963 & \textbf{0.9304} \\
\hline
\multicolumn{11}{c}{\textit{Text Sensitivity}} \\
\hline
Qwen2.5-VL-7B & 0.9980 & 0.9914 & 0.9620 & 0.9253 & \textbf{0.1384} & 0.9997 & 0.9987 & 0.9832 & 0.9669 & \textbf{0.1062} \\
Qwen2.5-VL-32B & 0.9982 & 0.9956 & 0.9575 & 0.9183 & \textbf{0.0875} & 0.9997 & 0.9990 & 0.9804 & 0.9655 & \textbf{0.1118} \\
LLaVA-v1.5 & 0.9973 & 0.9999 & 0.9778 & 0.9273 & \textbf{-0.0903} & 0.9994 & 1.0000 & 0.9837 & 0.9530 & \textbf{0.0105} \\
Phi-3.5-Vision & 0.9993 & 0.9949 & 0.9636 & 0.9461 & \textbf{0.2401} & 0.9997 & 0.9980 & 0.9794 & 0.9711 & \textbf{0.2789} \\
DeepEncoder & 0.9982 & 0.9931 & 0.9637 & 0.9352 & \textbf{0.2705} & 0.9996 & 0.9975 & 0.9819 & 0.9700 & \textbf{0.3894} \\
\hline
\end{tabular}
}
\captionof{table}{Complete sensitivity benchmark across aggregation strategies.}
\label{tab:full_pooling_results}

\end{strip}

\section{Results}
\subsection{The Blindness of Single-vector Aggregation in All Sensitivity Tests}
As shown in Table \ref{tab:full_pooling_results}, both aggregation methods, Mean Pooling and Max Pooling fail completely in both FinQA and TAT-DQA. The similarity scores stay near 1.0 in all our tests. No matter whether we change the text or the numbers, these methods cannot distinguish between the original and the altered document. This confirms that single-vector aggregation strategies are effectively blind to fine-grained semantic changes. In a dense retrieval index, when semantically different 
documents (e.g., \$1.2M vs. \$7.2M) have similarity scores 
near 1.0, they occupy the same vector space region. It becomes extremely difficult for 
queries to reliably rank one above the other, making 
precise version discrimination highly challenging in 
financial applications.

\subsection{Slight Recovery via Late Interaction}
Late-interaction methods such as MaxSim and MeanPatch operate at the patch level, which should theoretically preserve local details. However, as shown in Table~\ref{tab:full_pooling_results}, these methods provide only marginal improvement: MaxSim and MeanPatch scores remain above 0.99 across all sensitivity 
tests. This proves that the global texture (background layout, grid lines, 
and headers) is strong enough to overpower even these local metrics.

\subsection{MinPatch Shows the Hidden Signal}
Unlike single-vector aggregation, MinPatch successfully distinguishes between small and large errors. In Table \ref{tab:full_pooling_results}, similarity scores drop to about 0.71 for micro-semantic changes but fall much further to 0.51 for macro-semantic changes. This proves that the encoder does see the difference. We see similar results in the text sensitivity test. MinPatch scores drop to 0.09 for Qwen2.5-VL-32B, and even become negative for LLaVA. These results confirm that the vision encoder sees the error clearly. It is the aggregation process that obscures the signal. Notably, DeepEncoder consistently 
shows the highest MinPatch scores (least sensitive), likely because its OCR-optimized encoder learns representations invariant to pixel-level variations, making its aggregated representation less sensitive to single-digit changes compared to general vision encoders.

\begin{center}
\resizebox{\columnwidth}{!}{
\centering
\begin{tabular}{l|cc}
\hline
& \multicolumn{2}{c}{\textbf{FinQA}} \\
\hline
\textbf{Sensitivity Analysis} & \textbf{Qwen3-VL} & \textbf{GME-Qwen2-}  \\
& \textbf{Embedding-8B} & \textbf{VL-7B-Instruct}\\
\hline
Micro-Semantic & 0.9992 & 0.9970 \\
Macro-Semantic & 0.9976 & 0.9906  \\
Text Sensitivity & 0.9799 & 0.9363 \\
\hline
\end{tabular}
}
\captionof{table}{Evaluation of retrieval-optimized 
embedding models on sensitivity benchmarks (FinQA).}
\label{tab:retrieval_emb_results_finqa}
\end{center}

\begin{center}
\resizebox{\columnwidth}{!}{
\centering
\begin{tabular}{l|cc}
\hline
& \multicolumn{2}{c}{\textbf{TAT-DQA}} \\
\hline
\textbf{Sensitivity Analysis} & \textbf{Qwen3-VL} & \textbf{GME-Qwen2-}  \\
& \textbf{Embedding-8B} & \textbf{VL-7B-Instruct}\\
\hline
Micro-Semantic & 0.9999 & 0.9991 \\
Macro-Semantic & 0.9997 & 0.9906  \\
Text Sensitivity & 0.9932 & 0.9091 \\
\hline
\end{tabular}
}
\captionof{table}{Evaluation of retrieval-optimized 
embedding models on sensitivity benchmarks (TAT-DQA).}
\label{tab:retrieval_emb_results_tatdqa}
\end{center}

\subsection{Retrieval-Optimized Embedding Models}
A natural question is whether embedding models specifically trained for retrieval tasks can overcome the aggregation failure. As shown in Table~\ref{tab:retrieval_emb_results_finqa} and Table~\ref{tab:retrieval_emb_results_tatdqa}, both Qwen3-VL-Embedding-8B and GME-Qwen2-VL-7B-Instruct exhibit 
the same blindness (similarity near 1.0). This confirms that the 
failure is inherent to single-vector representations, regardless 
of training objectives.

\subsection{Preliminary Mitigation Attempts}
To explore whether simple aggregation strategies can mitigate global texture
dominance, we tested three approaches: variance-weighted pooling (VarWgt), attention-guided pooling (AttnGd), and top-k patch removal 
(TopK-R). As shown in Table~\ref{tab:mitigation_results}, all three 
methods provide only marginal improvement, with similarity scores remaining above 0.99 across all sensitivity tests. This confirms the severity of global texture dominance: background features are so pervasive that simple re-weighting or filtering strategies cannot recover the fine-grained signals. Combined with our MinPatch results (Table~\ref{tab:full_pooling_results}), which show similarity as low as 0.51, this demonstrates that 
the problem is fundamental to single-vector aggregation.

\subsection{Layout vs Tabular Data}
\label{sec:bias}
Since we want to know why single-vector aggregation fails on the sensitivity test, we perform a visual attention analysis to see whether the vision embeddings focus on the background layout (including non-informative headers) or the tabular data. The performance gap between the similarity scores of the background layout and the tabular data helps us identify what the single vector pays attention to. As shown in Table \ref{tab:visual_attention_results}, both 
Qwen2.5-VL-7B and Qwen2.5-VL-32B show a large bias 
($\Delta = +0.22$ and $+0.24$, respectively) in FinQA. This indicates the single vector focuses on the background layout rather than the tabular data. On the other hand, DeepSeek-DeepEncoder shows a lower bias ($\Delta = +0.08$) in FinQA. This suggests the single vector of DeepSeek focuses less on the background layout but still not enough to identify the changes. In TAT-DQA, the gap becomes negligible ($\Delta < 0.05$) due to global texture dominance. This prevents the single vector 
from distinguishing between table content and background layout, 
effectively blinding the single-vector representation to the 
actual data within visually complex tables.


\section{Conclusion}
In this paper, we develop a diagnostic benchmark to analyze the reliability of single-vector aggregation in visual document retrieval for table-centric financial documents. We find that although VLMs' vision encoders successfully capture fine-grained numeric and textual details, aggregation methods such as mean pooling obscure these signals due to global texture dominance. Our findings are robust across varying model scales, specialized retrieval-optimized architectures, and multiple mitigation strategies, all of which fail to recover the lost signals. These results highlight significant risks for single-vector 
visual document retrieval in financial applications and suggest that multi-vector retrieval approaches or learned aggregation methods are necessary for practical deployment.

\section*{Limitations}
Our diagnostic benchmark focuses on table-centric financial documents from two datasets (FinQA and TAT-DQA). While these represent common financial document formats, other types such as invoices, balance sheets with different layouts, or 
handwritten financial notes may exhibit different levels of global texture dominance. Extending the benchmark to cover a broader range of financial document types would strengthen the generalizability of our findings. Additionally, the findings on global texture dominance may not transfer to other domains such as natural images, which have fundamentally different visual characteristics (see Appendix~\ref{sec:sanity_check}). Furthermore, our benchmark may not cover all real-world semantic changes since we only work on specific numeric 
and textual perturbations. Future work should explore more diverse perturbation types and larger-scale evaluation across financial document categories. While our study focuses on 
diagnostic similarity analysis between document pairs, a full retrieval evaluation with ranking metrics (e.g., Recall@k, nDCG) over a large corpus would further contextualize the practical impact and is left for future work.

\section*{Acknowledgments}
This work was conducted as part of a HKUST Frontier Technology Research for Joint Institutes with Industry (FTRIS) project and was supported by WeBank under Grant No. WEB25BM01.

\bibliography{custom}

@inproceedings{zeiler2014visualizing,
  title={Visualizing and understanding convolutional networks},
  author={Zeiler, Matthew D and Fergus, Rob},
  booktitle={European conference on computer vision},
  pages={818--833},
  year={2014},
  organization={Springer}
}

@article{faysse2024colpali,
  title={Colpali: Efficient document retrieval with vision language models},
  author={Faysse, Manuel and Sibille, Hugues and Wu, Tony and Omrani, Bilel and Viaud, Gautier and Hudelot, C{\'e}line and Colombo, Pierre},
  journal={arXiv preprint arXiv:2407.01449},
  year={2024}
}

@misc{bai2025qwen25vltechnicalreport,
      title={Qwen2.5-VL Technical Report}, 
      author={Shuai Bai and Keqin Chen and Xuejing Liu and Jialin Wang and Wenbin Ge and Sibo Song and Kai Dang and Peng Wang and Shijie Wang and Jun Tang and Humen Zhong and Yuanzhi Zhu and Mingkun Yang and Zhaohai Li and Jianqiang Wan and Pengfei Wang and Wei Ding and Zheren Fu and Yiheng Xu and Jiabo Ye and Xi Zhang and Tianbao Xie and Zesen Cheng and Hang Zhang and Zhibo Yang and Haiyang Xu and Junyang Lin},
      year={2025},
      eprint={2502.13923},
      archivePrefix={arXiv},
      primaryClass={cs.CV},
      url={https://arxiv.org/abs/2502.13923}, 
}

@inproceedings{10.5555/3666122.3667638,
author = {Liu, Haotian and Li, Chunyuan and Wu, Qingyang and Lee, Yong Jae},
title = {Visual instruction tuning},
year = {2023},
publisher = {Curran Associates Inc.},
address = {Red Hook, NY, USA},
booktitle = {Proceedings of the 37th International Conference on Neural Information Processing Systems},
articleno = {1516},
numpages = {25},
location = {New Orleans, LA, USA},
series = {NIPS '23}
}

@inproceedings{chen-etal-2021-finqa,
    title = "{F}in{QA}: A Dataset of Numerical Reasoning over Financial Data",
    author = "Chen, Zhiyu  and
      Chen, Wenhu  and
      Smiley, Charese  and
      Shah, Sameena  and
      Borova, Iana  and
      Langdon, Dylan  and
      Moussa, Reema  and
      Beane, Matt  and
      Huang, Ting-Hao  and
      Routledge, Bryan  and
      Wang, William Yang",
    editor = "Moens, Marie-Francine  and
      Huang, Xuanjing  and
      Specia, Lucia  and
      Yih, Scott Wen-tau",
    booktitle = "Proceedings of the 2021 Conference on Empirical Methods in Natural Language Processing",
    month = nov,
    year = "2021",
    address = "Online and Punta Cana, Dominican Republic",
    publisher = "Association for Computational Linguistics",
    url = "https://aclanthology.org/2021.emnlp-main.300/",
    doi = "10.18653/v1/2021.emnlp-main.300",
    pages = "3697--3711"
}

@inproceedings{Zhu_2022, series={MM ’22},
   title={Towards Complex Document Understanding By Discrete Reasoning},
   url={http://dx.doi.org/10.1145/3503161.3548422},
   DOI={10.1145/3503161.3548422},
   booktitle={Proceedings of the 30th ACM International Conference on Multimedia},
   publisher={ACM},
   author={Zhu, Fengbin and Lei, Wenqiang and Feng, Fuli and Wang, Chao and Zhang, Haozhou and Chua, Tat-Seng},
   year={2022},
   month=oct, pages={4857–4866},
   collection={MM ’22} }

@inproceedings{10.1145/3397271.3401075,
author = {Khattab, Omar and Zaharia, Matei},
title = {ColBERT: Efficient and Effective Passage Search via Contextualized Late Interaction over BERT},
year = {2020},
isbn = {9781450380164},
publisher = {Association for Computing Machinery},
address = {New York, NY, USA},
url = {https://doi.org/10.1145/3397271.3401075},
doi = {10.1145/3397271.3401075},
booktitle = {Proceedings of the 43rd International ACM SIGIR Conference on Research and Development in Information Retrieval},
pages = {39–48},
numpages = {10},
location = {Virtual Event, China},
series = {SIGIR '20}
}

@misc{abdin2024phi3technicalreporthighly,
      title={Phi-3 Technical Report: A Highly Capable Language Model Locally on Your Phone}, 
      author={Marah Abdin and Jyoti Aneja and Hany Awadalla and Ahmed Awadallah and Ammar Ahmad Awan and Nguyen Bach and Amit Bahree and Arash Bakhtiari and Jianmin Bao and Harkirat Behl and Alon Benhaim and Misha Bilenko and Johan Bjorck and Sébastien Bubeck and Martin Cai and Qin Cai and Vishrav Chaudhary and Dong Chen and Dongdong Chen and Weizhu Chen and Yen-Chun Chen and Yi-Ling Chen and Hao Cheng and Parul Chopra and Xiyang Dai and Matthew Dixon and Ronen Eldan and Victor Fragoso and Jianfeng Gao and Mei Gao and Min Gao and Amit Garg and Allie Del Giorno and Abhishek Goswami and Suriya Gunasekar and Emman Haider and Junheng Hao and Russell J. Hewett and Wenxiang Hu and Jamie Huynh and Dan Iter and Sam Ade Jacobs and Mojan Javaheripi and Xin Jin and Nikos Karampatziakis and Piero Kauffmann and Mahoud Khademi and Dongwoo Kim and Young Jin Kim and Lev Kurilenko and James R. Lee and Yin Tat Lee and Yuanzhi Li and Yunsheng Li and Chen Liang and Lars Liden and Xihui Lin and Zeqi Lin and Ce Liu and Liyuan Liu and Mengchen Liu and Weishung Liu and Xiaodong Liu and Chong Luo and Piyush Madan and Ali Mahmoudzadeh and David Majercak and Matt Mazzola and Caio César Teodoro Mendes and Arindam Mitra and Hardik Modi and Anh Nguyen and Brandon Norick and Barun Patra and Daniel Perez-Becker and Thomas Portet and Reid Pryzant and Heyang Qin and Marko Radmilac and Liliang Ren and Gustavo de Rosa and Corby Rosset and Sambudha Roy and Olatunji Ruwase and Olli Saarikivi and Amin Saied and Adil Salim and Michael Santacroce and Shital Shah and Ning Shang and Hiteshi Sharma and Yelong Shen and Swadheen Shukla and Xia Song and Masahiro Tanaka and Andrea Tupini and Praneetha Vaddamanu and Chunyu Wang and Guanhua Wang and Lijuan Wang and Shuohang Wang and Xin Wang and Yu Wang and Rachel Ward and Wen Wen and Philipp Witte and Haiping Wu and Xiaoxia Wu and Michael Wyatt and Bin Xiao and Can Xu and Jiahang Xu and Weijian Xu and Jilong Xue and Sonali Yadav and Fan Yang and Jianwei Yang and Yifan Yang and Ziyi Yang and Donghan Yu and Lu Yuan and Chenruidong Zhang and Cyril Zhang and Jianwen Zhang and Li Lyna Zhang and Yi Zhang and Yue Zhang and Yunan Zhang and Xiren Zhou},
      year={2024},
      eprint={2404.14219},
      archivePrefix={arXiv},
      primaryClass={cs.CL},
      url={https://arxiv.org/abs/2404.14219}, 
}

@misc{wei2025deepseekocrcontextsopticalcompression,
      title={DeepSeek-OCR: Contexts Optical Compression}, 
      author={Haoran Wei and Yaofeng Sun and Yukun Li},
      year={2025},
      eprint={2510.18234},
      archivePrefix={arXiv},
      primaryClass={cs.CV},
      url={https://arxiv.org/abs/2510.18234}, 
}

@inproceedings{li-etal-2025-lost-embeddings,
    title = "Lost in Embeddings: Information Loss in Vision{--}Language Models",
    author = "Li, Wenyan  and
      Tang, Raphael  and
      Li, Chengzu  and
      Zhang, Caiqi  and
      Vuli{\'c}, Ivan  and
      S{\o}gaard, Anders",
    editor = "Christodoulopoulos, Christos  and
      Chakraborty, Tanmoy  and
      Rose, Carolyn  and
      Peng, Violet",
    booktitle = "Findings of the Association for Computational Linguistics: EMNLP 2025",
    month = nov,
    year = "2025",
    address = "Suzhou, China",
    publisher = "Association for Computational Linguistics",
    url = "https://aclanthology.org/2025.findings-emnlp.1235/",
    doi = "10.18653/v1/2025.findings-emnlp.1235",
    pages = "22676--22693",
    ISBN = "979-8-89176-335-7"
}

@inproceedings{zhu-etal-2021-tat,
    title = "{TAT}-{QA}: A Question Answering Benchmark on a Hybrid of Tabular and Textual Content in Finance",
    author = "Zhu, Fengbin  and
      Lei, Wenqiang  and
      Huang, Youcheng  and
      Wang, Chao  and
      Zhang, Shuo  and
      Lv, Jiancheng  and
      Feng, Fuli  and
      Chua, Tat-Seng",
    editor = "Zong, Chengqing  and
      Xia, Fei  and
      Li, Wenjie  and
      Navigli, Roberto",
    booktitle = "Proceedings of the 59th Annual Meeting of the Association for Computational Linguistics and the 11th International Joint Conference on Natural Language Processing (Volume 1: Long Papers)",
    month = aug,
    year = "2021",
    address = "Online",
    publisher = "Association for Computational Linguistics",
    url = "https://aclanthology.org/2021.acl-long.254/",
    doi = "10.18653/v1/2021.acl-long.254",
    pages = "3277--3287"
}

@inproceedings{wolf-etal-2020-transformers,
    title = "Transformers: State-of-the-Art Natural Language Processing",
    author = "Wolf, Thomas  and
      Debut, Lysandre  and
      Sanh, Victor  and
      Chaumond, Julien  and
      Delangue, Clement  and
      Moi, Anthony  and
      Cistac, Pierric  and
      Rault, Tim  and
      Louf, Remi  and
      Funtowicz, Morgan  and
      Davison, Joe  and
      Shleifer, Sam  and
      von Platen, Patrick  and
      Ma, Clara  and
      Jernite, Yacine  and
      Plu, Julien  and
      Xu, Canwen  and
      Le Scao, Teven  and
      Gugger, Sylvain  and
      Drame, Mariama  and
      Lhoest, Quentin  and
      Rush, Alexander",
    editor = "Liu, Qun  and
      Schlangen, David",
    booktitle = "Proceedings of the 2020 Conference on Empirical Methods in Natural Language Processing: System Demonstrations",
    month = oct,
    year = "2020",
    address = "Online",
    publisher = "Association for Computational Linguistics",
    url = "https://aclanthology.org/2020.emnlp-demos.6/",
    doi = "10.18653/v1/2020.emnlp-demos.6",
    pages = "38--45"
}

@misc{dadopoulos2025metadatadrivenretrievalaugmentedgenerationfinancial,
      title={Metadata-Driven Retrieval-Augmented Generation for Financial Question Answering}, 
      author={Michail Dadopoulos and Anestis Ladas and Stratos Moschidis and Ioannis Negkakis},
      year={2025},
      eprint={2510.24402},
      archivePrefix={arXiv},
      primaryClass={cs.IR},
      url={https://arxiv.org/abs/2510.24402}, 
}

@misc{kim2025optimizingretrievalstrategiesfinancial,
      title={Optimizing Retrieval Strategies for Financial Question Answering Documents in Retrieval-Augmented Generation Systems}, 
      author={Sejong Kim and Hyunseo Song and Hyunwoo Seo and Hyunjun Kim},
      year={2025},
      eprint={2503.15191},
      archivePrefix={arXiv},
      primaryClass={cs.IR},
      url={https://arxiv.org/abs/2503.15191}, 
}

@misc{si2025tabragtabulardocumentretrieval,
      title={TabRAG: Tabular Document Retrieval via Structured Language Representations}, 
      author={Jacob Si and Mike Qu and Michelle Lee and Yingzhen Li},
      year={2025},
      eprint={2511.06582},
      archivePrefix={arXiv},
      primaryClass={cs.CL},
      url={https://arxiv.org/abs/2511.06582}, 
}

@inproceedings{yu-etal-2025-tablerag,
    title = "{T}able{RAG}: A Retrieval Augmented Generation Framework for Heterogeneous Document Reasoning",
    author = "Yu, Xiaohan  and
      Jian, Pu  and
      Chen, Chong",
    editor = "Christodoulopoulos, Christos  and
      Chakraborty, Tanmoy  and
      Rose, Carolyn  and
      Peng, Violet",
    booktitle = "Proceedings of the 2025 Conference on Empirical Methods in Natural Language Processing",
    month = nov,
    year = "2025",
    address = "Suzhou, China",
    publisher = "Association for Computational Linguistics",
    url = "https://aclanthology.org/2025.emnlp-main.710/",
    doi = "10.18653/v1/2025.emnlp-main.710",
    pages = "14063--14082",
    ISBN = "979-8-89176-332-6"
}

@inproceedings{ma-etal-2024-unifying,
    title = "Unifying Multimodal Retrieval via Document Screenshot Embedding",
    author = "Ma, Xueguang  and
      Lin, Sheng-Chieh  and
      Li, Minghan  and
      Chen, Wenhu  and
      Lin, Jimmy",
    editor = "Al-Onaizan, Yaser  and
      Bansal, Mohit  and
      Chen, Yun-Nung",
    booktitle = "Proceedings of the 2024 Conference on Empirical Methods in Natural Language Processing",
    month = nov,
    year = "2024",
    address = "Miami, Florida, USA",
    publisher = "Association for Computational Linguistics",
    url = "https://aclanthology.org/2024.emnlp-main.373/",
    doi = "10.18653/v1/2024.emnlp-main.373",
    pages = "6492--6505"
}

@inproceedings{10.1007/978-3-031-88714-7_22,
author = {MacAvaney, Sean and Mallia, Antonio and Tonellotto, Nicola},
title = {Efficient Constant-Space Multi-vector Retrieval},
year = {2025},
isbn = {978-3-031-88713-0},
publisher = {Springer-Verlag},
address = {Berlin, Heidelberg},
url = {https://doi.org/10.1007/978-3-031-88714-7_22},
doi = {10.1007/978-3-031-88714-7_22},
booktitle = {Advances in Information Retrieval: 47th European Conference on Information Retrieval, ECIR 2025, Lucca, Italy, April 6–10, 2025, Proceedings, Part III},
pages = {237–245},
numpages = {9},
location = {Lucca, Italy}
}

@inproceedings{10.5555/3495724.3496517,
author = {Lewis, Patrick and Perez, Ethan and Piktus, Aleksandra and Petroni, Fabio and Karpukhin, Vladimir and Goyal, Naman and K\"{u}ttler, Heinrich and Lewis, Mike and Yih, Wen-tau and Rockt\"{a}schel, Tim and Riedel, Sebastian and Kiela, Douwe},
title = {Retrieval-augmented generation for knowledge-intensive NLP tasks},
year = {2020},
isbn = {9781713829546},
publisher = {Curran Associates Inc.},
address = {Red Hook, NY, USA},
booktitle = {Proceedings of the 34th International Conference on Neural Information Processing Systems},
articleno = {793},
numpages = {16},
location = {Vancouver, BC, Canada},
series = {NIPS '20}
}

@inproceedings{
yu2025visrag,
title={Vis{RAG}: Vision-based Retrieval-augmented Generation on Multi-modality Documents},
author={Shi Yu and Chaoyue Tang and Bokai Xu and Junbo Cui and Junhao Ran and Yukun Yan and Zhenghao Liu and Shuo Wang and Xu Han and Zhiyuan Liu and Maosong Sun},
booktitle={The Thirteenth International Conference on Learning Representations},
year={2025},
url={https://openreview.net/forum?id=zG459X3Xge}
}

@inproceedings{kim2022donut,
  title     = {OCR-Free Document Understanding Transformer},
  author    = {Kim, Geewook and Hong, Teakgyu and Yim, Moonbin and Nam, JeongYeon and Park, Jinyoung and Yim, Jinyeong and Hwang, Wonseok and Yun, Sangdoo and Han, Dongyoon and Park, Seunghyun},
  booktitle = {European Conference on Computer Vision (ECCV)},
  year      = {2022}
}

@misc{zhang2024gme,
      title={GME: Improving Universal Multimodal Retrieval by Multimodal LLMs}, 
      author={Zhang, Xin and Zhang, Yanzhao and Xie, Wen and Li, Mingxin and Dai, Ziqi and Long, Dingkun and Xie, Pengjun and Zhang, Meishan and Li, Wenjie and Zhang, Min},
      year={2024},
      eprint={2412.16855},
      archivePrefix={arXiv},
      primaryClass={cs.CL},
      url={http://arxiv.org/abs/2412.16855}, 
}

@article{qwen3vlembedding,
  title={Qwen3-VL-Embedding and Qwen3-VL-Reranker: A Unified Framework for State-of-the-Art Multimodal Retrieval and Ranking},
  author={Li, Mingxin and Zhang, Yanzhao and Long, Dingkun and Chen Keqin and Song, Sibo and Bai, Shuai and Yang, Zhibo and Xie, Pengjun and Yang, An and Liu, Dayiheng and Zhou, Jingren and Lin, Junyang},
  journal={arXiv preprint arXiv:2601.04720},
  year={2026}
}
\clearpage 
\onecolumn 
\appendix

\section{Dataset Details}
\label{sec:dataset_details}

\paragraph{FinQA.} FinQA \cite{chen-etal-2021-finqa} focuses on numerical reasoning in financial reports. We create the image set by manually taking screenshots of the test documents\footnote{\url{https://finqasite.github.io/explore.html}} to simulate real-world viewing conditions.

\paragraph{TAT-DQA.} TAT-DQA~\cite{Zhu_2022}, an extension of the TAT-QA dataset~\cite{zhu-etal-2021-tat}, is considered harder than FinQA. It contains multi-page documents with dense financial tables. We extract these images directly from the source PDFs.

\paragraph{Masking Procedure.} Unlike automated bounding boxes which often suffer from localization errors, we manually verify the visual boundaries of the table region for each document. This ensures pixel-perfect alignment with the table's semantic content, avoiding partial occlusions or leftover artifacts common in automated methods.

\section{Implementation Details}
\label{sec:appendix_implementation}

\paragraph{Setup.}
We implemented all models using the HuggingFace Transformers library~\citep{wolf-etal-2020-transformers} on a single NVIDIA RTX 5880 Ada Generation GPU (48GB). To ensure a fair comparison, we utilized the official pre-trained weights for all architectures (e.g., \texttt{Qwen/Qwen2.5-VL-7B-Instruct}). For DeepEncoder, we adapted the implementation from \url{https://github.com/Volkopat/VLM-Optical-Encoder}. Embedding similarity was computed using \textbf{Cosine Similarity}, and all input images were resized to the model's default resolution to prevent resizing artifacts.

\newpage
\section{Mitigation Strategy Details}
\label{sec:mitigation_details}

To explore whether simple modifications to the aggregation 
process can preserve fine-grained signals, we design three 
straightforward baseline strategies. We emphasize that these strategies are not proposed as optimal solutions, but rather as simple probes to test whether straightforward aggregation modifications can recover fine-grained signals lost during single-vector compression.

\begin{itemize}
    \item \textbf{Variance-Weighted Pooling (VarWgt):} 
    Assigns higher weights to patches with greater variance 
    across embedding dimensions, under the hypothesis that 
    informative patches exhibit higher variance than 
    repetitive background patches.
    
    \item \textbf{Attention-Guided Pooling (AttnGd):} 
    Computes a patch-level self-similarity matrix via normalized dot products among patch embeddings, and uses the standard deviation of each patch’s similarity scores as a proxy for patch informativeness. Patches with low diversity are down-weighted under the hypothesis that uniformly similar patches correspond to repetitive background regions.
    
    \item \textbf{Top-k Removal (TopK-R):} 
    Computes cosine similarity between spatially aligned patch pairs from the original and perturbed documents, removes the top-$k$ most similar aligned patches, and mean-pools the remaining patches. We use a fixed $k=50$ as a simple heuristic. Note that this corresponds to different removal proportions across encoders (around 9–19\% depending on model).
\end{itemize}

\begin{table}[h]
\centering
\resizebox{\textwidth}{!}{
\begin{tabular}{l|ccc|ccc}
\hline
& \multicolumn{3}{c|}{\textbf{FinQA}} & \multicolumn{3}{c}{\textbf{TAT-DQA}} \\
\hline
\textbf{Model} & \textbf{VarWgt} & \textbf{AttnGd} & \textbf{TopK-R}  & \textbf{VarWgt} & \textbf{AttnGd} & \textbf{TopK-R} \\
\hline
\multicolumn{7}{c}{\textit{Micro-Semantic Sensitivity (Values close to 1.0 indicate blindness)}} \\
\hline
Qwen2.5-VL-7B & 0.9999 & 1.0000 & 1.0000 & 1.0000 & 1.0000 & 1.0000   \\
Qwen2.5-VL-32B & 0.9997 & 0.9999 & 0.9999 &  1.0000 & 1.0000 & 1.0000   \\
LLaVA-v1.5 & 1.0000 & 1.0000 & 1.0000 &  1.0000 & 1.0000 & 1.0000  \\
Phi-3.5-Vision & 1.0000 & 1.0000 & 1.0000 & 1.0000 & 1.0000 & 1.0000  \\
DeepEncoder & 0.9999 & 0.9999 & 0.9999 & 1.0000 & 1.0000 & 1.0000  \\
\hline
\multicolumn{7}{c}{\textit{Macro-Semantic Sensitivity}} \\
\hline
Qwen2.5-VL-7B & 0.9997 & 0.9998 & 0.9998 & 0.9996 & 0.9999 & 0.9999  \\
Qwen2.5-VL-32B & 0.9997 & 0.9998 & 0.9998 & 0.9999 & 1.0000 & 1.0000  \\
LLaVA-v1.5 & 1.0000 & 0.9999 & 0.9999 & 0.9999 & 0.9999 & 0.9999 \\
Phi-3.5-Vision & 0.9999 & 0.9999 & 0.9999 & 0.9999 & 1.0000 & 1.0000  \\
DeepEncoder & 0.9994 & 0.9995 & 0.9994 & 0.9998 & 0.9997 & 0.9996 \\
\hline
\multicolumn{7}{c}{\textit{Text Sensitivity}} \\
\hline
Qwen2.5-VL-7B & 0.9962 & 0.9982 & 0.9979 & 0.9992 & 0.9997 & 0.9832 \\
Qwen2.5-VL-32B & 0.9972 & 0.9980 & 0.9980 & 0.9996 & 0.9996 & 0.9996 \\
LLaVA-v1.5 & 0.9998 & 0.9974 & 0.9981 &  0.9999 & 0.9992 & 0.9992 \\
Phi-3.5-Vision & 0.9989 & 0.9990 & 0.9988 & 0.9995  & 0.9996 & 0.9995 \\
DeepEncoder & 0.9980 & 0.9982 & 0.9978 & 0.9996 & 0.9996 & 0.9995 \\
\hline
\end{tabular}
}
\caption{Evaluation of mitigation strategies on sensitivity benchmarks.}
\label{tab:mitigation_results}
\end{table}

\newpage
\section{Visual Attention Analysis}
\begin{table}[h] 
\small
\centering
\begin{tabular}{l|ccc|ccc}
\hline
& \multicolumn{3}{c|}{\textbf{FinQA}} & \multicolumn{3}{c}{\textbf{TAT-DQA}} \\
\hline
\textbf{Model} & \textbf{Sim to Data} & \textbf{Sim to Layout} & \textbf{Gap} & \textbf{Sim to Data } & \textbf{Sim to Layout} & \textbf{Gap}\\
& (Table Only) & & & (Table Only) & & \\
\hline
\multicolumn{7}{c}{\textit{STRATEGY: MEAN POOLING (Values close to 1.0 indicate blindness)}} \\
\hline
Qwen2.5-VL-7B & 0.7548 & 0.9791 & \textbf{0.2243} & 0.9268 & 0.9464 & \textbf{0.0196} \\
Qwen2.5-VL-32B & 0.7369 & 0.9724 & \textbf{0.2355} & 0.9343 & 0.9222 & \textbf{-0.0121} \\
LLaVA-v1.5 & 0.8040 & 0.9847 & \textbf{0.1807} & 0.9364 & 0.9684 & \textbf{0.0320} \\
Phi-3.5-Vision & 0.7171 & 0.9911 & \textbf{0.2740} & 0.9190 & 0.9693 & \textbf{0.0503} \\
DeepEncoder & 0.9106 & 0.9866 & \textbf{0.0760} & 0.9222	& 0.9667 &	\textbf{0.0446}  \\
\hline
\multicolumn{7}{c}{\textit{STRATEGY: MAX POOLING}} \\
\hline
Qwen2.5-VL-7B & 0.9019 & 0.9677 & 0.0658 & 0.9710 & 0.9766 & 0.0056 \\
Qwen2.5-VL-32B & 0.9641 & 0.9822 & 0.0181 & 0.9859 & 0.9809 & -0.0051 \\
LLaVA-v1.5 & 0.9986 & 0.9997 & 0.0012 & 0.9995 & 0.9997 & 0.0002\\
Phi-3.5-Vision & 0.9208 & 0.9854 & 0.0646 & 0.9629 & 0.9781 & 0.0152\\
DeepEncoder & 0.9626 & 0.9817 & 0.0190 & 0.9762 & 0.9811 &	0.0050\\
\hline
\multicolumn{7}{c}{\textit{STRATEGY: LATE INTERACT (MAXSIM)}} \\
\hline
Qwen2.5-VL-7B & 0.6062 & 0.8948 & 0.2886 & 0.7948 & 0.8286 & 0.0338  \\
Qwen2.5-VL-32B & 0.5401 & 0.8821 & 0.3420 & 0.7570 & 0.7954 & 0.0384 \\
LLaVA-v1.5 & 0.6935 & 0.9418 & 0.2482 & 0.8095 & 0.8810 & 0.0715 \\
Phi-3.5-Vision & 0.6109 & 0.9129 & 0.3020 & 0.7431 & 0.8297 & 0.0866 \\
DeepEncoder & 0.7584 & 0.8944 & 0.1360 & 0.7995 & 0.8584 & 0.0588 \\
\hline
\multicolumn{7}{c}{\textit{STRATEGY: LATE INTERACT (MEANPATCH)}} \\
\hline
Qwen2.5-VL-7B & 0.3841 & 0.8231 & \textbf{0.4390} & 0.6171 & 0.6974 & \textbf{0.0803}  \\
Qwen2.5-VL-32B & 0.3364 & 0.8007 & \textbf{0.4643} & 0.6081 & 0.6672 & \textbf{0.0591}  \\
LLaVA-v1.5 & 0.3788 & 0.8346 & \textbf{0.4558} & 0.5177 & 0.6764 & \textbf{0.1587} \\
Phi-3.5-Vision & 0.3028 & 0.8706 & \textbf{0.5678} & 0.5480 & 0.7150 & \textbf{0.1670} \\
DeepEncoder & 0.6232 & 0.8309 & \textbf{0.2077} & 0.6367 & 0.7398 & \textbf{0.1031} \\
\hline
\multicolumn{7}{c}{\textit{STRATEGY: LATE INTERACT (MINPATCH)}} \\
\hline
Qwen2.5-VL-7B & -0.1575 & 0.0083 & 0.1658 & -0.0455 & 0.0079 & 0.0534  \\
Qwen2.5-VL-32B & -0.0080 & 0.0407 & 0.0487 & 0.0183 & 0.0272 & 0.0089  \\
LLaVA-v1.5 & -0.1257 & -0.0592 & 0.0666 & -0.1335 & -0.1360 & -0.0025 \\
Phi-3.5-Vision & -0.1617 & 0.0494 & 0.2111 & -0.1091 & -0.0526 & 0.0566 \\
DeepEncoder & 0.1670 & 0.1727 & 0.0058 & 0.1798 & 0.1844 & 0.0046 \\
\hline
\end{tabular}
\caption{Complete visual attention analysis results 
across aggregation strategies.}
\label{tab:visual_attention_results}
\end{table}

\section{Sanity Check — Natural Images vs. Financial Documents}
\label{sec:sanity_check}
\begin{table}[h]
\centering
\begin{tabular}{l|c|c}
\hline
\textbf{Model} & \textbf{Natural Image (Cat vs Dog)} & \textbf{Financial Doc (\$1.2 vs \$1.3)}  \\
\hline
Qwen3-VL-Embedding-8B & 0.1192 & 0.9992  \\
GME-Qwen2-VL-7B-Instruct & 0.2165 & 0.9970  \\
Qwen2.5-VL-7B & 0.4038 & 0.9999 \\
Qwen2.5-VL-32B & 0.5117 & 0.9999  \\
\hline
\end{tabular}
\caption{Domain comparison: natural images vs. financial documents.}
\label{tab:domain_comparision_results}
\end{table}

To confirm that the aggregation failure is domain-specific, we compared natural images (cat vs. dog) with financial documents (\$1.2M vs. \$1.3M) using the same models. Results show natural images maintain discriminability (similarity 0.12--0.51), while financial documents collapse (similarity near 1.0), confirming a domain gap of 0.5--0.9. This validates our hypothesis that sparse numeric signals in background-dominated documents are uniquely vulnerable to single-vector aggregation.

\end{document}